\documentclass[runningheads]{llncs}

\usepackage[T1]{fontenc}
\usepackage{lmodern}
\usepackage{graphicx}
\usepackage{amsmath,amssymb} 
\usepackage{color}
\usepackage{booktabs}
\usepackage{multirow}
\usepackage[width=122mm,left=12mm,paperwidth=146mm,height=193mm,top=12mm,paperheight=217mm]{geometry}
\usepackage{hyperref}
\hypersetup{
    colorlinks=true, 
    urlcolor=black, 
    linkcolor=black, 
    citecolor=black 
}

\newcommand{\numcircled}[1]{\raisebox{.5pt}{\textcircled{\raisebox{-.9pt} {#1}}}}
\newcommand{\citep}{\cite}
\newcommand{\citet}{\cite}

\begin{document}
\pagestyle{headings}
\mainmatter
\def\ECCV18SubNumber{1653}  

\title{Multi-Task Learning by Deep Collaboration and Application in Facial Landmark Detection}


\author{
    Ludovic Trottier \qquad Philippe Gigu\`ere \qquad Brahim Chaib-draa \\
    {\footnotesize \url{ludovic.trottier.1@ulaval.ca}} \\
    {\footnotesize \{\url{philippe.giguere, brahim.chaib-draa}\}\url{@ift.ulaval.ca}}
}
\institute{Laval University, Qu\'ebec, Canada}

\titlerunning{Trottier et al.}
\authorrunning{Multi-Task Learning by Deep Collaboration}

\maketitle

\begin{abstract}
    Convolutional neural networks (CNNs) have become the most successful approach in many vision-related domains. However, they are limited to domains where data is abundant. 
    Recent works have looked at multi-task learning (MTL) to mitigate data scarcity by leveraging domain-specific information from related tasks.
    In this paper, we present a novel soft-parameter sharing mechanism for CNNs in a MTL setting, which we refer to as Deep Collaboration. We propose taking into account the notion that task relevance depends on depth by using lateral transformation blocs with skip connections. This allows extracting task-specific features at various depth without sacrificing features relevant to all tasks.
    We show that CNNs connected with our Deep Collaboration obtain better accuracy on facial landmark detection with related tasks. We finally verify that our approach effectively allows knowledge sharing by showing depth-specific influence of tasks that we know are related.
\end{abstract}

\section{Introduction}
\label{sec:intro}



Over the past few years, Convolutional Neural Networks (CNNs) have become the leading approach in many vision-related tasks~\citep{krizhevsky2012imagenet}. Their ability to learn a hierarchy of increasingly abstract concepts allows them to transform complex high-dimensional input images into simple low-dimensional output features. CNNs have been used in many settings, but their need to have a large amount of data during training has restricted them to domains where data is abundant. Optimizing CNNs is tricky not only because of problems like vanishing / exploding gradients~\citep{hochreiter1998vanishing}, but also because they typically have many parameters to be learned. While previous works have looked at supervised and unsupervised pre-training to improve generalization, others have considered casting their original single-task problem into a new Multi-Task Learning (MTL) problem~\citet{zhang2017survey}. As Caruana (1998)~\citet{caruana1998multitask} explained in his seminal work: ``MTL improves generalization by leveraging the domain-specific information contained in the training signals of related tasks". Exploring new ways to more efficiently gather information from related tasks would help to further improve generalization on the main one.

MTL has proven its value in several domains over the years. It has become a dominant field of machine learning~\citep{zhang2014review}, with many influential works~\citep{evgeniou2004regularized}. Although MTL dates back several years, recent major advances in Deep Learning (DL) opened up opportunities for novel contributions. Works on grasping~\citep{pinto2017learning}, pedestrian detection~\citep{tian2015pedestrian}, natural language processing~\citep{liu2015representation}, face recognition~\citep{yim2015rotating}\citep{yin2017multi} and object detection~\citep{misra2016cross} helped MTL make a resurgence in the DL community. They have shown the potential of MTL to mitigate data scarcity when training deep networks, which has influenced its growing popularity

MTL approaches can generally be divided into two major categories: \emph{hard} and \emph{soft} parameter sharing~\citep{DBLP:journals/corr/Ruder17a}. Hard-parameter sharing dates back to the original work of Caruana (1998) and is the most common of the two. Approaches in this category have a shared central section with many heads (one per task). Features from specific tasks compete together and those relevant to all tasks are favored. Recent works in DL have shown that hard-parameter sharing can be successful~\citep{ranjan2016hyperface}\citep{zhang2014facial}\citep{pinto2017learning}\citep{yin2017multi}. However, a too large emphasis on features relevant to all tasks can be harmful for learning high-level features specific to a particular task. These types of specific features are usually needed to obtain a good representation for the particular task. Also, shared layers are prone to be contaminated by noise coming from noxious tasks~\citep{liu2017adversarial}. These limitations can be detrimental even though hard-parameter sharing reduces the risk of over-fitting~\citep{baxter1997bayesian}.

Soft-parameter sharing has been proposed as an alternative to alleviate these drawbacks. Approaches in this category substitute the shared central section by separate task-specific CNNs, but provide a knowledge sharing mechanism to connect them. Each CNN can then learn task-specific features and share their knowledge without interfering with others. Recent works in this category have looked at regularizing the distance between task-specific parameters with a $\ell_2$ norm~\citep{duong2015low} or a trace norm~\citep{yang2016trace}, training shared and private LSTM submodules~\citep{liu2017adversarial}, partitioning the hidden layers into subspaces~\citep{ruder2017sluice} and regularizing the FC layers with tensor normal priors~\citep{long2015learning}. In the domain of continual learning, progressive network~\citep{rusu2016progressive} has also shown promising results for sequential transfer learning, by employing lateral connections to previously learned networks.

In this paper, we present a novel soft-parameter knowledge sharing mechanism for connecting task-specific CNNs in a MTL framework. We refer to our approach as Deep Collaboration. We define connectivity in terms of a \textit{collaborative block} that uses two non-linear transformations with lateral connections. One aggregates task-specific features into global features, and the other merges back the global features into each task-specific CNN. Our collaborative block is differentiable and can be dropped in any existing CNN architectures as a whole. We evaluated our approach on the problem of facial landmark detection in a MTL framework and obtained better results in comparison to other approaches of the literature. We further assess the objectivity of our training framework by randomly varying the contribution of each related tasks. Finally, we verify that our collaborative block enables knowledge sharing with an ablation study that shows the depth-specific influence of tasks that we know are related.

The content of our paper is organized as follows. In Section~\ref{sec:related-work}, we present related works in MTL and facial landmark detection. We elaborate on our approach in Section~\ref{sec:deep-collaboration}, and present experimental results in Section~\ref{sec:experiments}. We finally conclude our paper in Section~\ref{sec:conclusion}. Our code is available here:~\citep{trottier2018DeepCollaborationGithub}.

\section{Related Work}
\label{sec:related-work}

\subsection{Multi-Task Learning}
\label{ssec:related-work-mtl}

Our proposed Deep Collaboration knowledge sharing mechanism is related to other existing approaches. One is Cross-Stitch (XS)~\citep{misra2016cross}, which connects task-specific CNNs by \emph{linearly} combining their feature maps at certain depths. One drawback of XS is that it is limited to capture only linear dependencies between each CNN. In contrast to XS, our approach uses \emph{non-linear} transformations in order to capture more complex dependencies. 

Another related approach is Tasks-Constrained Deep Convolutional Network (TCDCN)~\citep{zhang2014facial}. The authors proposed an early-stopping criterion to remove auxiliary tasks that start to overfit before becoming detrimental to the main task. This approach has however several hyper-parameters to be selected manually. For each task, it has an hyper-parameter controlling the period length of the local window and a threshold that stops the task when the criterion exceeds it. Unlike TCDCN, our approach has no hyper-parameters that need to be tuned to the tasks at hand. Our collaborative block consists of a series of Batch Normalization~\citep{ioffe2015batch}, ReLU~\citep{nair2010rectified}, and convolutional layers shaped in a standard setting that is commonly found in nowadays works. 

Our proposed approach is also related to HyperFace~\citep{ranjan2016hyperface}. The authors proposed to fuse the layers at various depth and exploit features of different levels of complexity. Their goal was to allow low-level features with better localization properties to help tasks such as landmark localization and pose detection, and allow high-level features with better class-specific properties to help tasks like face detection and gender recognition. Although HyperFace is in the hard-parameter sharing category and is not entirely related to our approach, the idea of \textit{feature fusion} is also central in our work. Instead of fusing the features at intermediate layers of a single CNN, our approach aggregates same-level features of multiple CNNs, at different depth independently.

\subsection{Facial Landmark Detection}
\label{ssec:related-work-fld}

Facial Landmark Detection (FLD) is an essential component in many face-related tasks~\citep{sun2013deep}\citep{zhang2016joint}\citep{jourabloo2016large}\citep{baltruvsaitis2016openface}. FLD can be described as follows: given the image of a face of a person, the goal is to predict the $(x,y)$-position of specific landmarks associated with key features of the visage. Applications such as face recognition~\citep{ding2015robust}, face validation~\citep{taigman2014deepface}, facial feature detection and tacking~\citep{zhang2014improving} rely on the ability to correctly find the location of these distinct facial landmarks in order to succeed. Localizing facial key points like the center of the eyes, the corners of the mouth, the tip of the nose and the earlobes is however a challenging problem when many lighting conditions, head poses, facial expressions and occlusions increase diversity of the face images. In addition to integrating this variability into the estimation process, a FLD model must also take into account a number of correlated factors. For instance, although both an angry person and a sad person have frowned eyebrows, an angry person will have pinched lips while a sad person will have sunken mouth corners~\citep{fabian2016emotionet}. A particularity of datasets geared towards FLD is that they are particularly well-suited for MTL. In addition to containing the position of the facial landmarks, these datasets also contain a number of other labels that can be used to defined auxiliary tasks. Gender recognition, smile recognition, glasses recognition or face orientation are examples of tasks often chosen to evaluate MTL approaches.

\section{Deep Collaboration}
\label{sec:deep-collaboration}

Given $T$ task-specific Convolutional Neural Networks (CNNs), our goal is to connect them with lateral connections in order to allow domain-specific information sharing. We define connectivity in terms of a \textit{collaborative block} containing two distinct non-linear transformations. One aggregates task-specific features into global features, and the other merges back the global features into each task-specific CNN. Our collaborative block is differentiable and can be dropped in any existing CNN architectures as a whole. For this reason, we make no assumption about the structure of the task-specific CNNs. Our approach can even work with different CNNs, but for the sake of simplicity, we suppose that the CNNs are the same. We refer to it as the \textit{underlying network}.

We also decompose the underlying network as a series of blocks. Each block can be as small as a single layer, as large as the whole network itself, or based on simple rules, such as grouping all layers with matching spatial dimensions or grouping every $n$ subsequent layers. The arrangement of the layers into blocks does not change the composition of the underlying network. We only use it to make explicit the depth at which we connect the task-specific CNNs.

\begin{figure}[t]
    \centering
    \includegraphics[width=0.8\linewidth]{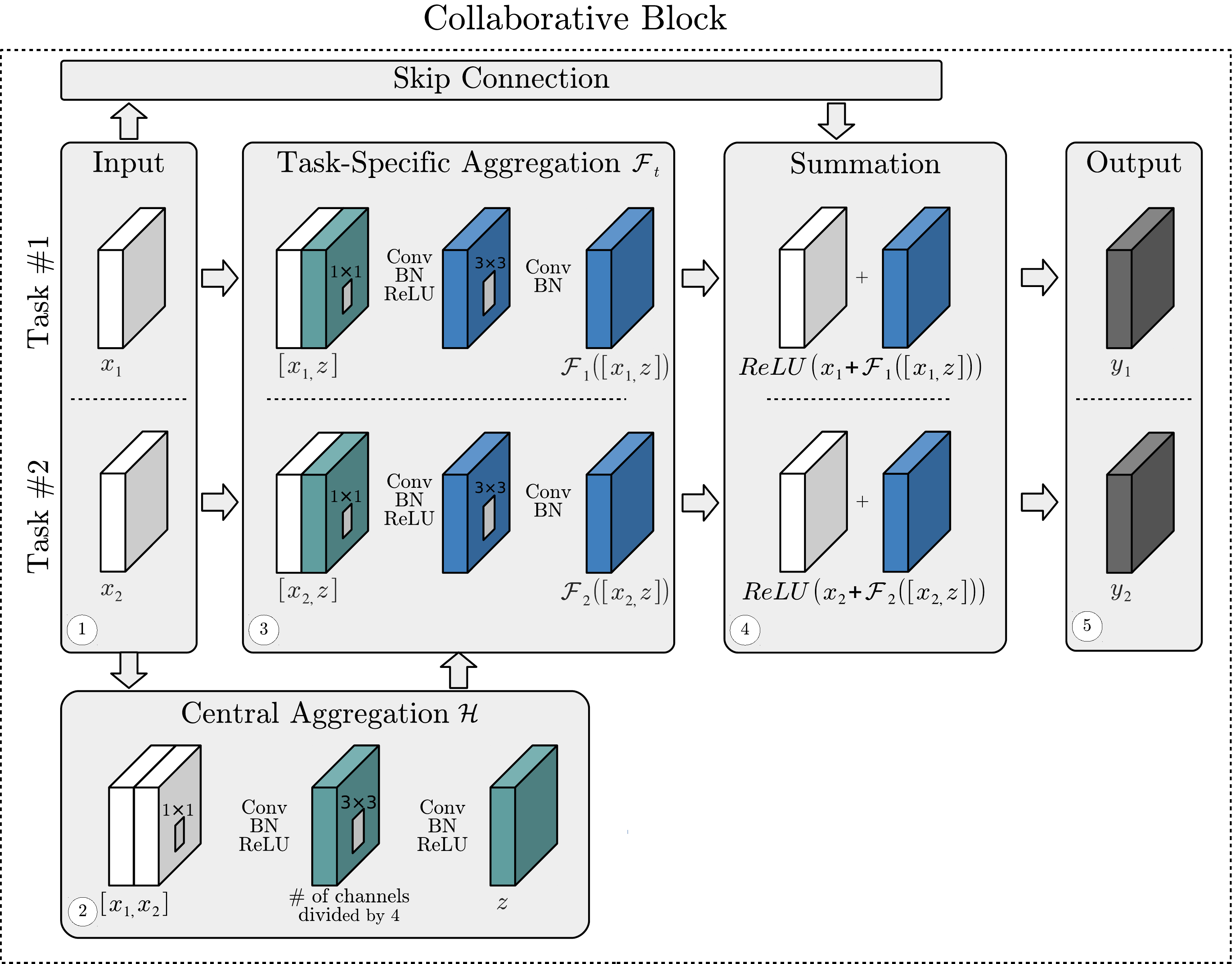}
    \caption{Example of our collaborative block applied on the feature maps of two task-specific networks. The input feature maps (shown in \numcircled{1}) are first concatenated depth-wise and transformed into a global feature map (\numcircled{2}). The global feature map is then concatenate with each input feature map individually and transformed into task-specific feature maps (\numcircled{3}). Each resulting feature map is then added back to the input feature map using a skip connection (\numcircled{4}), which gives the final outputs of the block (\numcircled{5}).}
    \label{fig:collaborative-block}
\end{figure}

Since our collaborative block can be inserted at any depth, we also drop the depth index on the feature maps to further simplify the equations. As such, we define the feature map output of a block at a certain depth as $x_t$, where $ t \in \{1 \ldots T\}$ is the task index. Our approach takes as input all task-specific feature maps $x_t$ and processes them into new feature maps $y_t$ as follows:
\begin{align}
\label{eq:collaborative-block}
z &= \mathcal{H}([x_1, \ldots, x_T]) \,, & y_t &= ReLU\left( x_t + \mathcal{F}_t([x_t, z]) \right) \, ,
\end{align}
where $\mathcal{H}$ and $\mathcal{F}_t$ represent the central and the task-specific aggregations respectively, and $[\cdot]$ denotes depth-wise concatenation. We refer to Eq.~\eqref{eq:collaborative-block} as our \textit{collaborative block}. The goal of $\mathcal{H}$ is to combine all task-specific feature maps $x_t$ into a global feature map $z$ representing unified knowledge, while the goal of $\mathcal{F}$ is to merge back the global feature map $z$ with each task-specific input $x_t$. The compositional structure of $\mathcal{H}$ and $\mathcal{F}$ is as follows:
\begin{align}
\mathcal{H}(\cdot) &= (ReLU \circ BN \circ Conv_{(3 \times 3)} \circ ReLU \circ BN \circ Conv_{(1 \times 1)})(\cdot) \, , \\
\mathcal{F}(\cdot) &= (BN \circ Conv_{(3 \times 3)} \circ ReLU  \circ BN \circ Conv_{(1 \times 1)})(\cdot) \, ,
\end{align}
where $BN$ stands for Batch Normalization~\citep{ioffe2015batch}, $Conv_{(h \times w)}$ for a standard convolutional layer with filters of size $(h \times w)$, and $\circ$ is the usual function composition. The first $Conv_{(1 \times 1)}$ layer in $\mathcal{H}$ divides the number of feature maps by a factor of $4$, while the first $Conv_{(1 \times 1)}$ layer in $\mathcal{F}$ divides it to match the size of $x_t$. An illustration of our collaborative block is shown in Fig.~\ref{fig:collaborative-block}.

One particularity of our approach is that we use a skip connection in mapping $\mathcal{F}$. Recent works~\citep{he2016identity}\citep{he2016deep}\citep{huang2016deep}\citep{xie2017aggregated}\citep{veit2016residual} have shown that networks with identity skip connections are more easily able to learn proper input-output mappings. Inspired by these works, we opted for an identity skip connection in $\mathcal{F}$ in order to more easily learn the proper mapping to integrate domain-specific information from the other tasks. In particular, identity skip connections put an incentive on learning the identity mapping. We can see this by the ease at which the network can obtain the identity mapping by simply pushing all the weights in $\mathcal{F}$ towards zero. In our MTL context, the identity mapping can be seen as a way to remove the influence of the global features $z$. This allows to take into account the cases where integrating $z$ back to the task-specific features $x_t$ would not help.

Another motivation for using an identity skip connection around the global feature map $z$ comes from the fact that depth influences the relevance of each task towards another. Some task-specific CNNs can benefit more when they share their low-level features than their high-level features, while other benefit more in the other way. For instance, tasks such as landmark localization and pose detection profit more from low-level features containing better localization properties, while tasks such as face detection and gender recognition profit more from class-specific high-level features. Considering that CNNs learn a hierarchy of increasingly abstract features, our collaborative block can take into account task relevance by deactivating a different set of residual mappings $\mathcal{F}_t$ based on the depth at which it is inserted. An example of such specialization will be shown in our ablative study in Section~\ref{ssec:task-relevance}.

\begin{figure}[t]
    \centering
    \includegraphics[width=\linewidth]{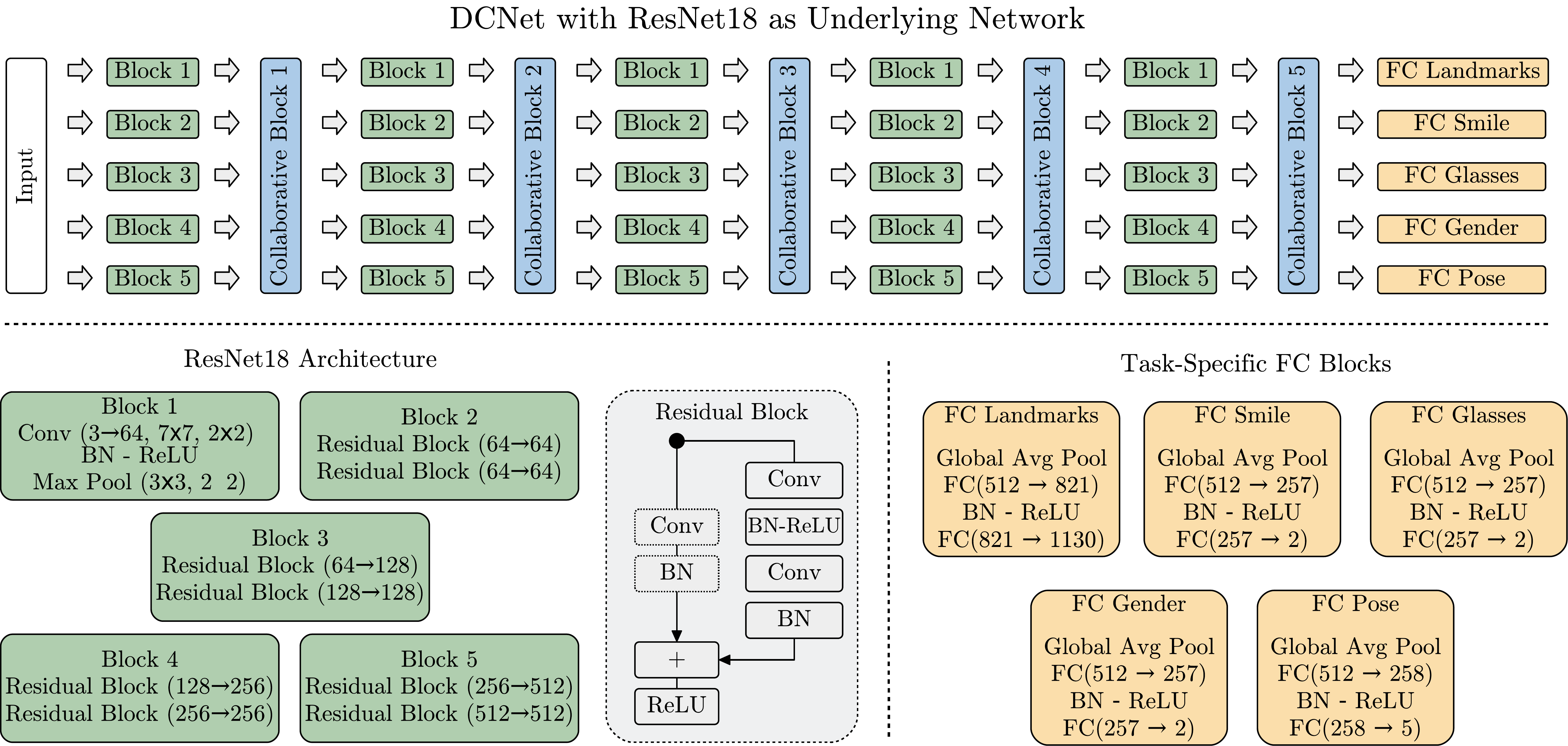}
    \caption{Deep Collaboration Network (DCNet) using ResNet18 as underlying network in a MTL setting on the MTFL dataset. The top part shows the block structure of ResNet18 interleaved with our proposed collaborative block, while the bottom part details each residual and task-specific FC blocks.}
    \label{fig:collaborative-resnet18}
\end{figure}

Fig.~\ref{fig:collaborative-resnet18} presents an example of inserting our collaborative block at different depths in a MTL framework on the MTFL dataset~\citep{zhang2014facial}. In this particular case, we opted for a ResNet18 as underlying network. We refer to this network as our Deep Collaboration Network (DCNet). As we can see in the top part of the figure, integrating our approach comes down to interleaving the underlying network block structure with our collaborative block. Each collaborative block receives as input the output of each task-specific block, processes them as detailed in Eq.~\eqref{eq:collaborative-block}, and sends the result back to each task-specific network. Adding our approach to any underlying network can be done by simply following the same pattern of interleaving the network block structure with our collaborative block.

\section{Experiments}
\label{sec:experiments}

In this section, we detail our Multi-Task Learning (MTL) training framework and present our experiments in Facial Landmark Detection (FLD) tasks. We further evaluate the effect of data scarcity on performance and illustrate an example of knowledge sharing between task-specific CNNs with an ablation study.

\subsection{Multi-Task Learning Training Framework}
\label{ssec:mtl-framework}

The goal of Facial Landmark Detection (FLD) is to predict the $(x,y)$-position of specific landmarks associated with key features of the visage. While the number and type of landmarks are specific to each dataset, examples of standard landmarks to be predicted are the corners of the mouth, the tip of the nose and the center of the eyes. In addition to the facial landmarks, each dataset further defines a number of related tasks. These related tasks also vary from one dataset to another, and are typically gender recognition, smile recognition, glasses recognition or face orientation.

On a more technical level, we define a learning framework in which we treat each task as a classification problem. While this is straightforward for gender, smile and glasses recognition as they are already classification tasks, it is a bit more tricky for face orientation and FLD. For face orientation, instead of predicting the roll, yaw and pitch real value as in a regression problem, we divide each component into 30 degrees wide bins and predict the label of the bin corresponding to the value. Similarly for FLD, rather than predicting the real $(x,y)$-position of each landmark, we divide the image into 1 pixel wide bins and predict the label of the bin corresponding to the value. Note that we still use the original real values when comparing our prediction with the ground truth, so that we incorporate our approximation errors in the final score.

We report our results using the landmark failure rate metric~\citep{zhang2014facial}, which is defined as follows: we first compute the mean distance between the predicted landmarks and the ground truth landmarks, then normalize it by the inter-ocular distance from the center of the eyes. A normalized mean distance greater than $10\%$ is reported as a failure.


\subsection{Facial Landmark Detection on the MTFL Task}
\label{ssec:mtfl-results}

\begin{figure}[t]
    \centering
    \begin{minipage}{0.49\linewidth}
        \includegraphics[width=\linewidth]{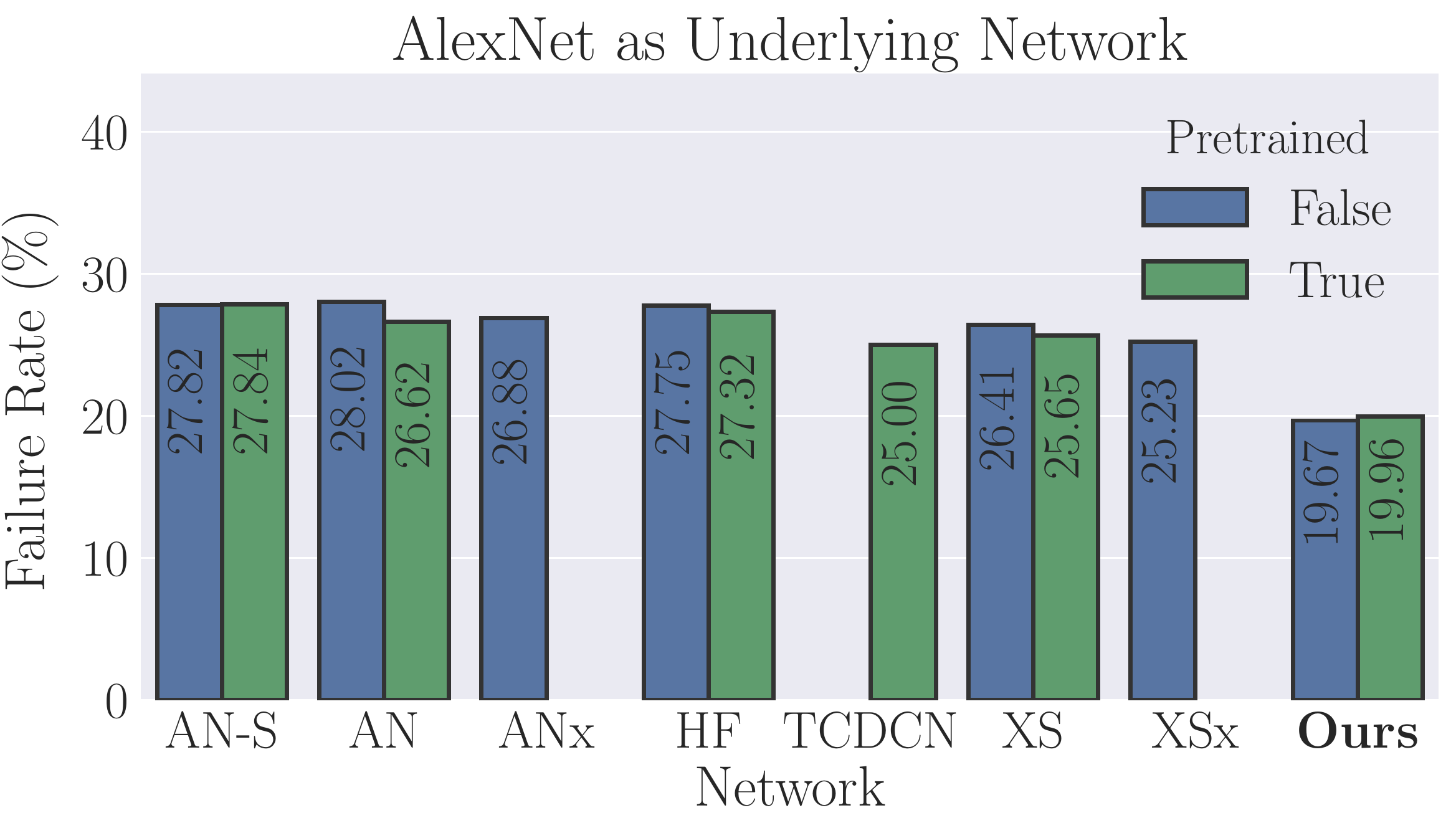}
    \end{minipage}
    \begin{minipage}{0.49\linewidth}
        \includegraphics[width=\linewidth]{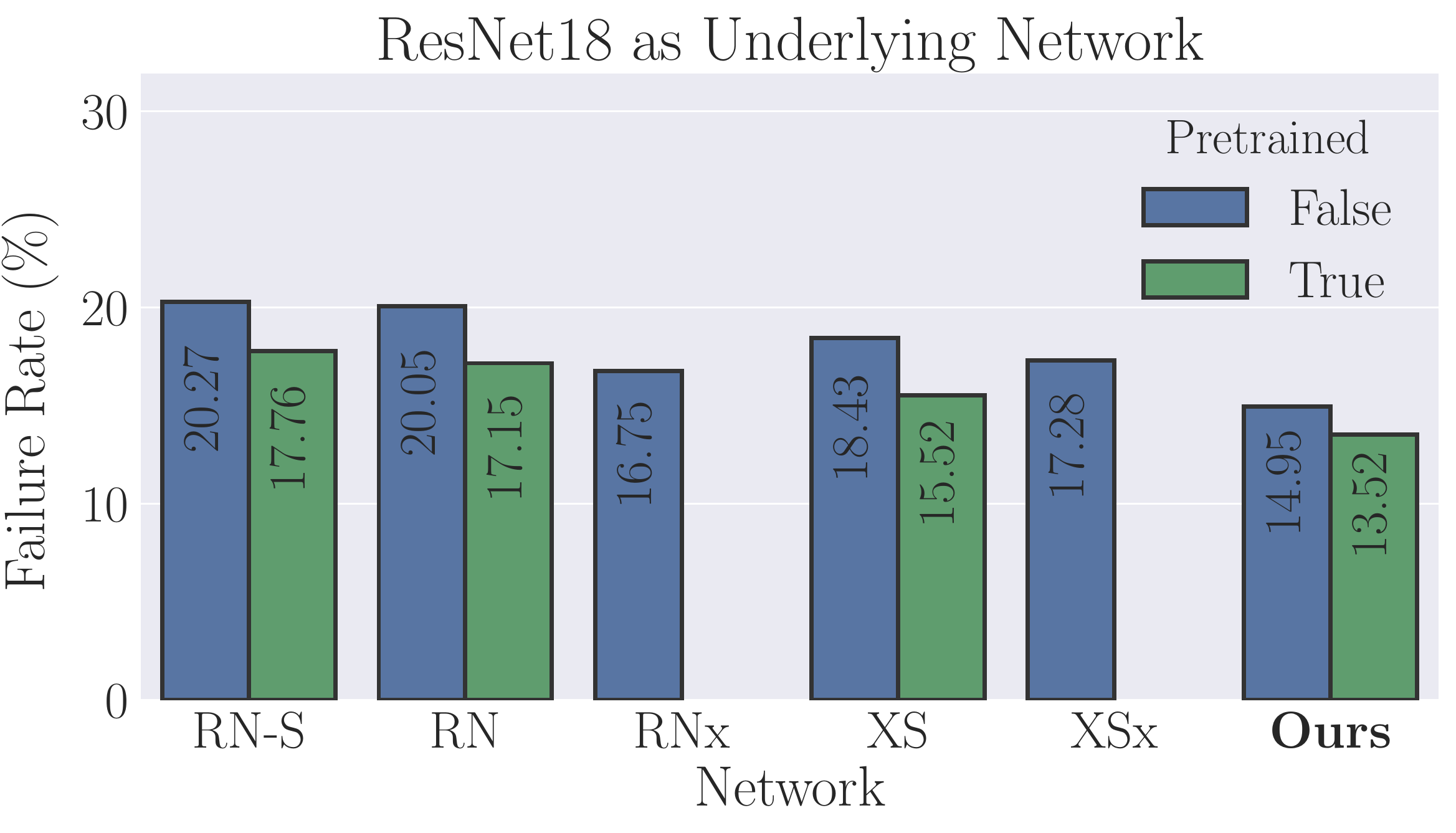}
    \end{minipage}
    \begin{minipage}{0.49\linewidth}
    \includegraphics[width=\linewidth]{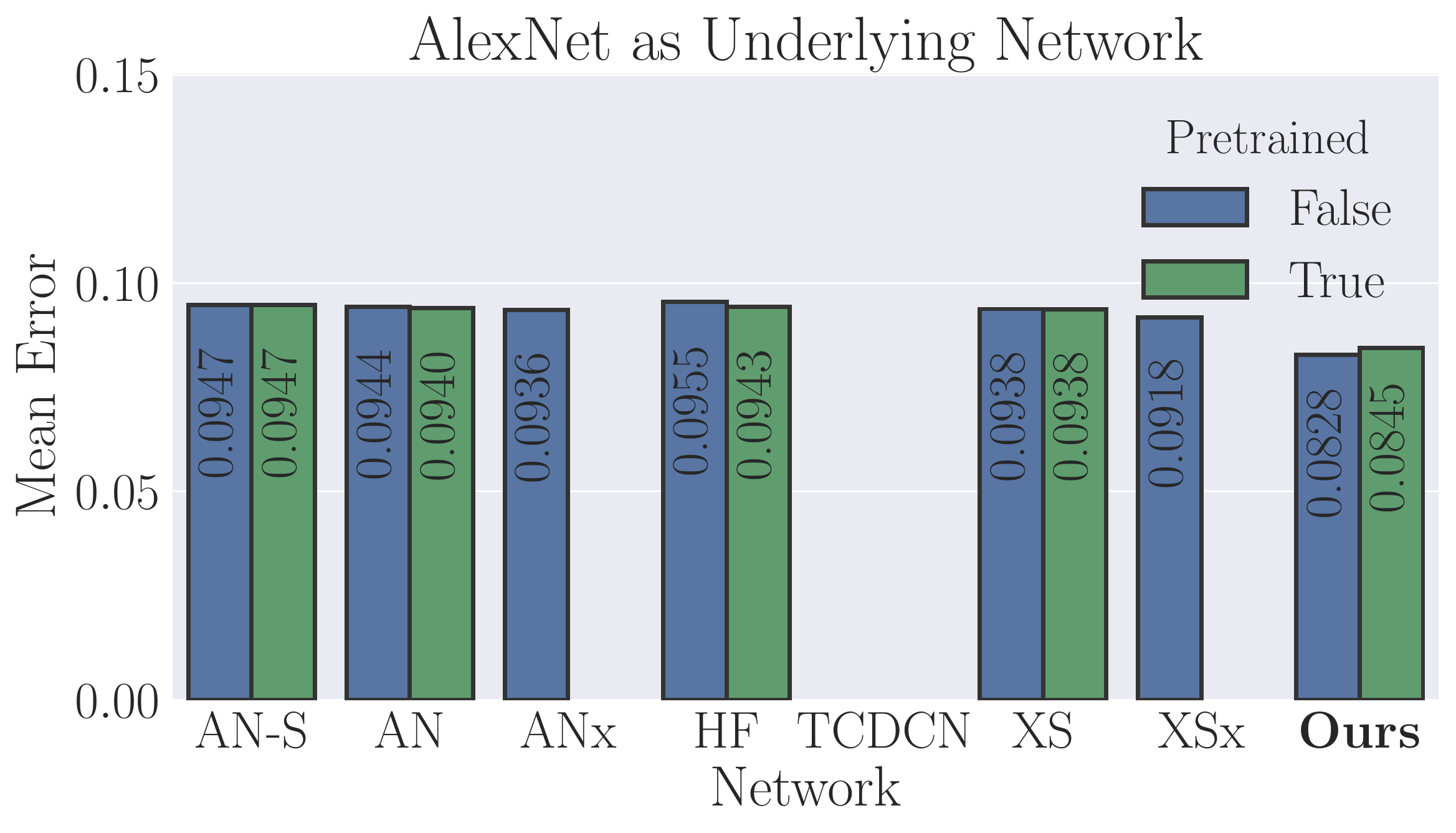}
    \end{minipage}
    \begin{minipage}{0.49\linewidth}
        \includegraphics[width=\linewidth]{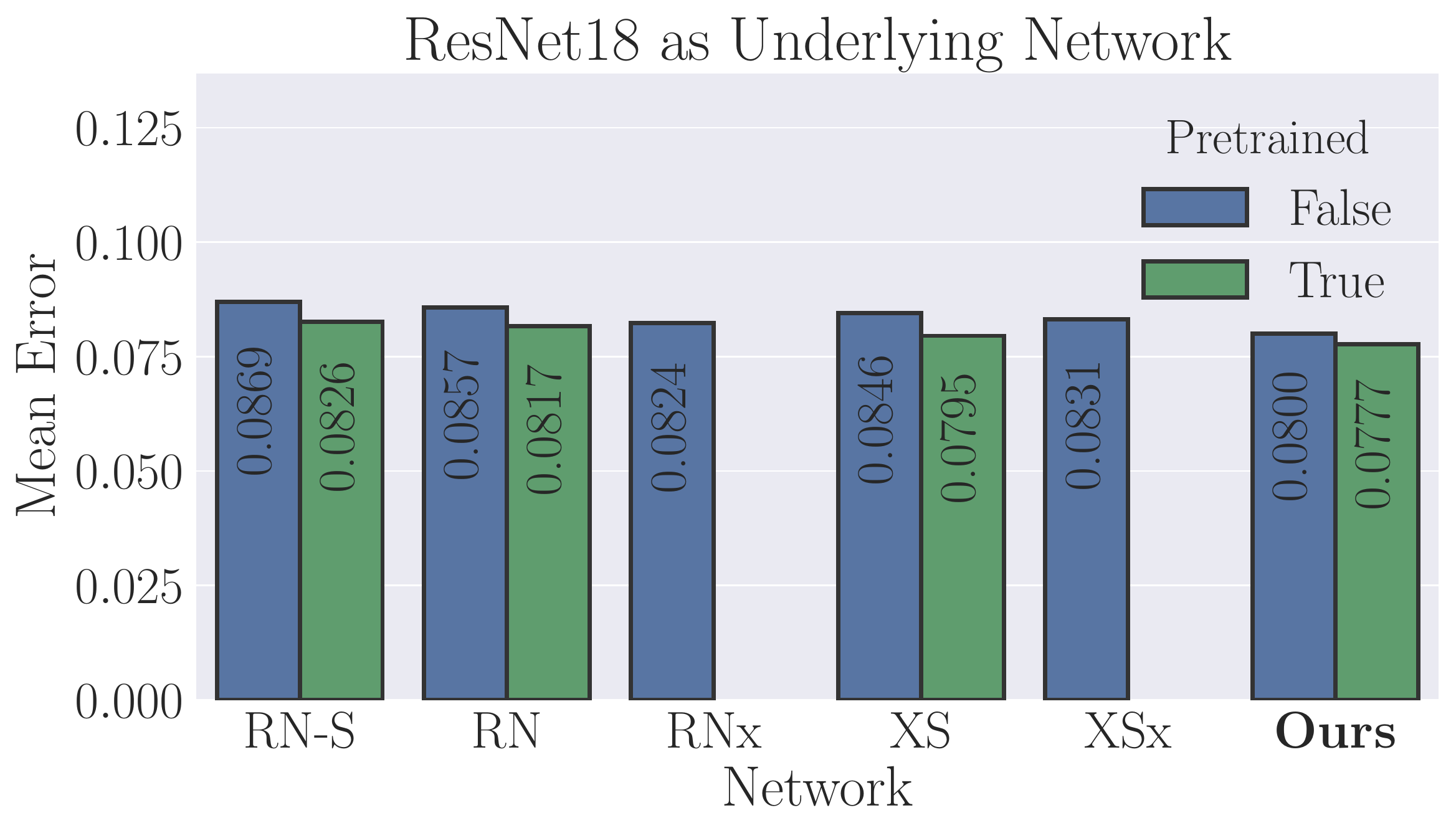}
    \end{minipage}
    \caption{Landmark failure rates (\%) on the MTFL task. The reported values are the average over the last five epochs, averaged over three tries. The left plot presents our results with AlexNet as the underlying network, while the right one with ResNet18. AN-S and RN-S stand for single-task training, AN and RN for multi-task training with a single central network, ANx and RNx for multi-task training with a single central network widen to match the number of parameters of our approach, HF for HyperFace, TCDCN for \citet{zhang2014facial}'s approach and XS for Cross-Stitch. In each instance, the left column (blue) is for un-pretrained networks, while the right column (green) is for pre-trained networks. Our proposed approach obtains the lowest failure rates overall.}
    \label{fig:results-mtfl}
\end{figure}

\begin{figure}[t]
    \begin{minipage}{\linewidth}
        \begin{minipage}{\linewidth}
            \centering
            MTFL
        \end{minipage}
        \begin{minipage}{0.24\linewidth}
            \includegraphics[width=\linewidth]{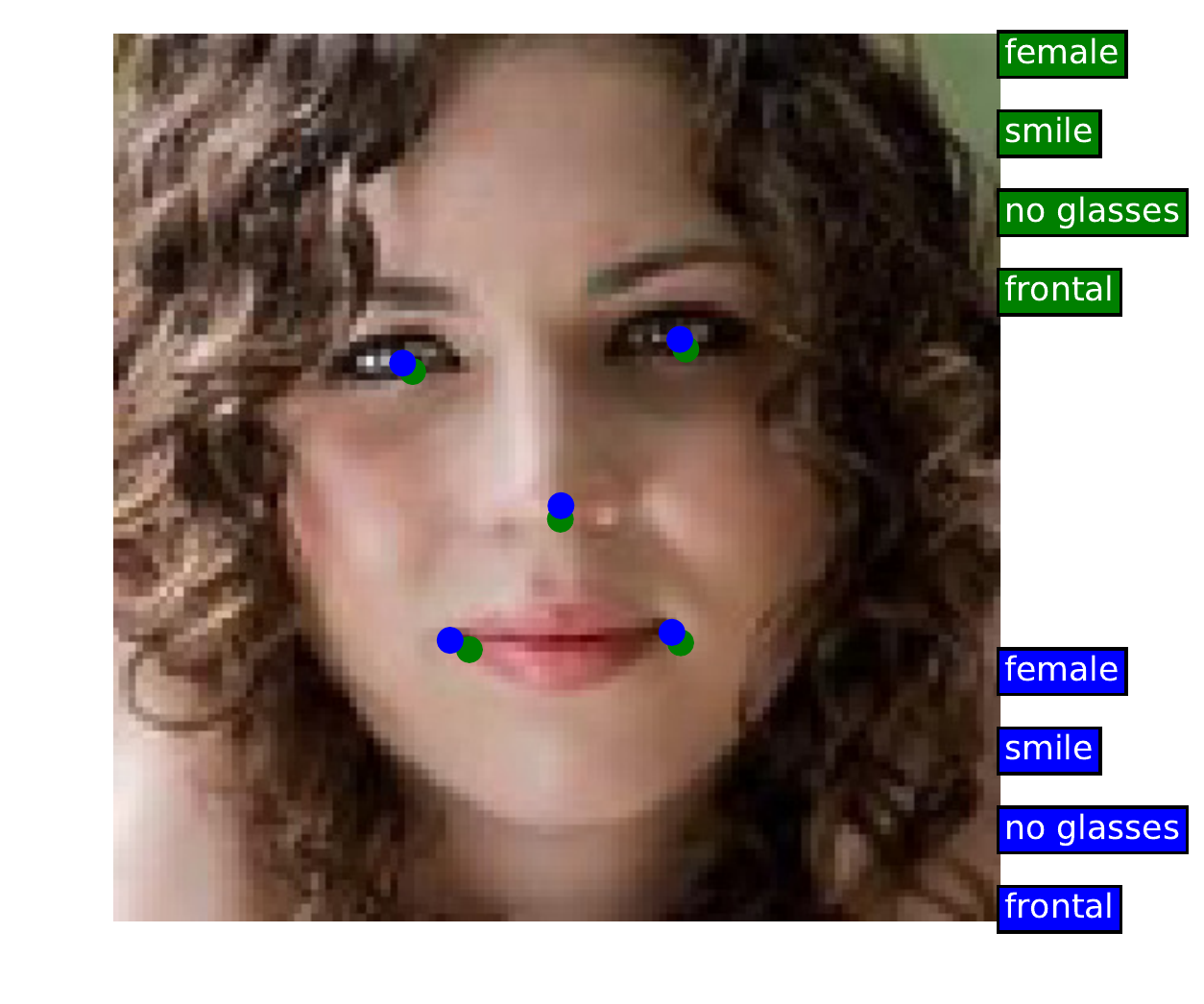}
        \end{minipage}
        \begin{minipage}{0.24\linewidth}
            \includegraphics[width=\linewidth]{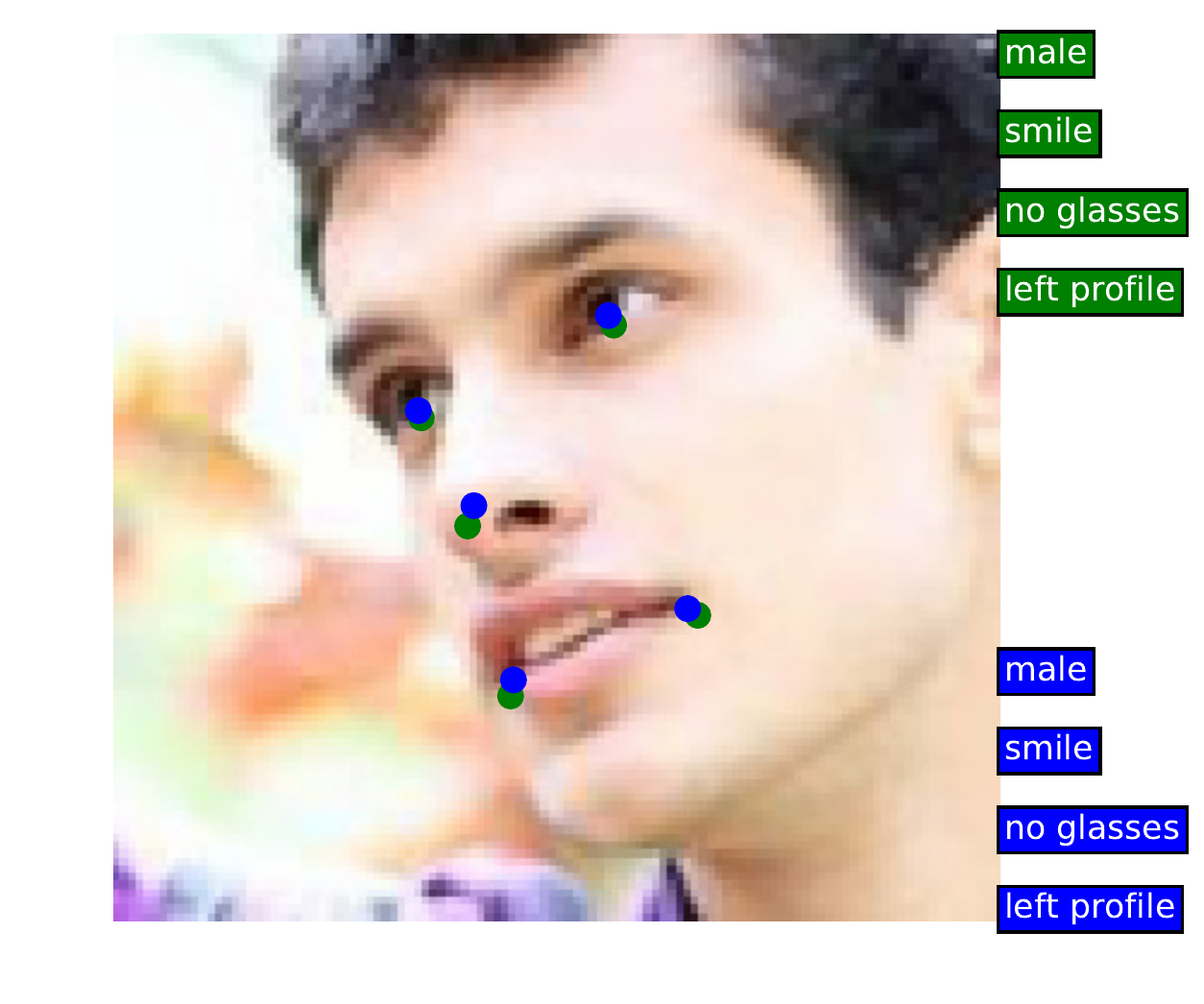}
        \end{minipage}
        \begin{minipage}{0.24\linewidth}
            \includegraphics[width=\linewidth]{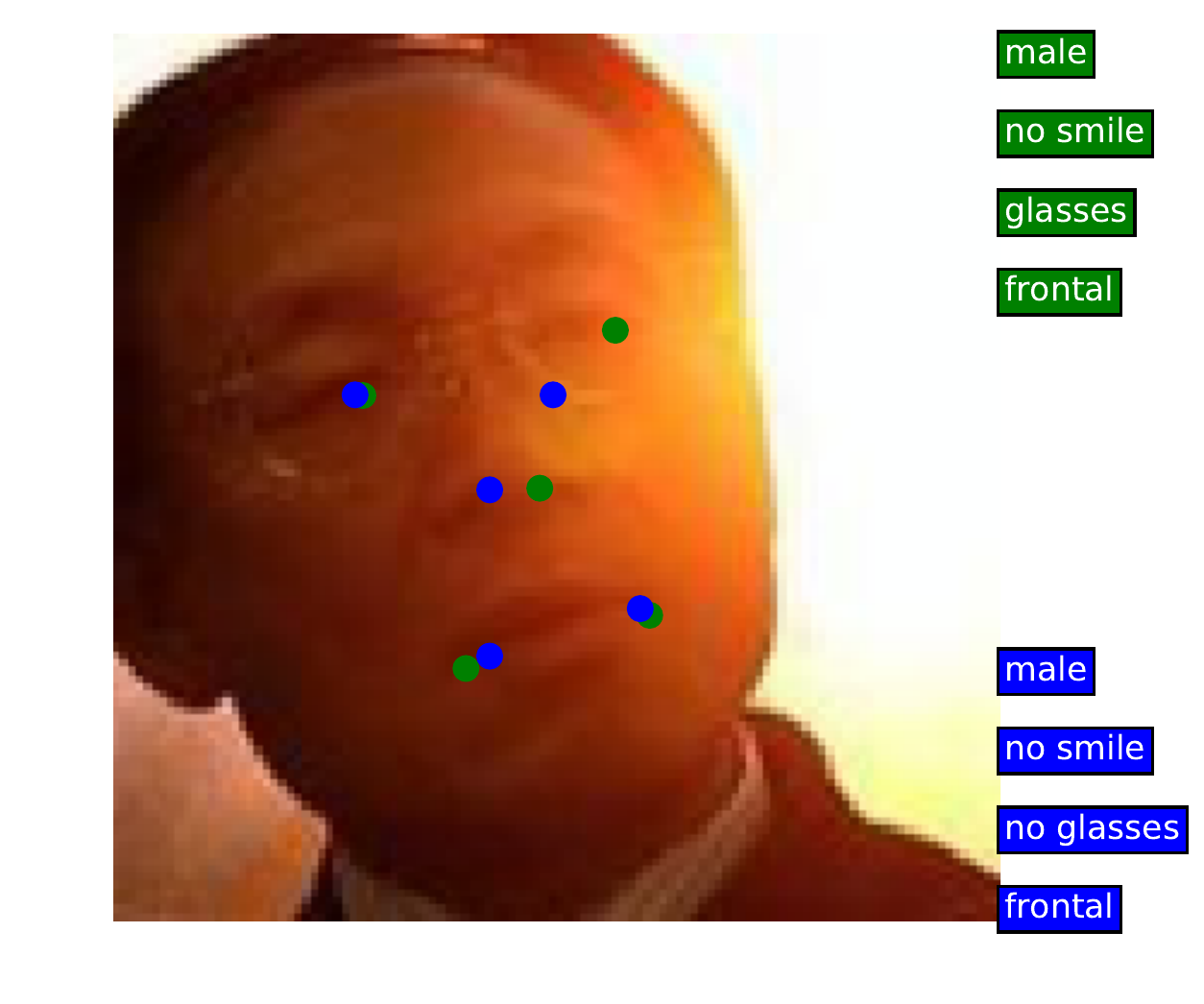}
        \end{minipage}
        \begin{minipage}{0.24\linewidth}
            \includegraphics[width=\linewidth]{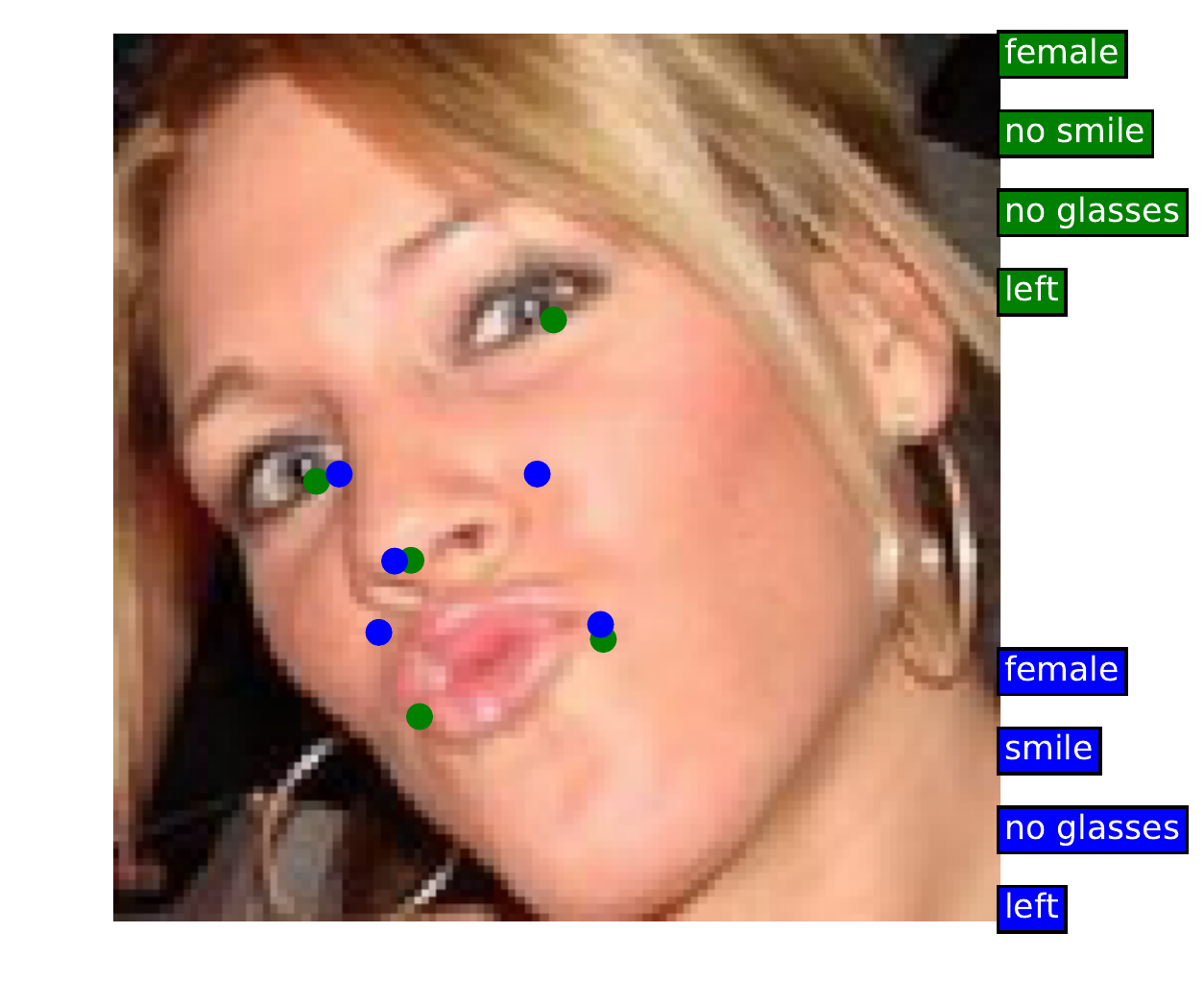}
        \end{minipage}
        \hrule
        \vspace{3pt}
        \begin{minipage}{\linewidth}
            \centering
            AFLW
        \end{minipage}        
        \begin{minipage}{0.24\linewidth}
            \includegraphics[width=\linewidth]{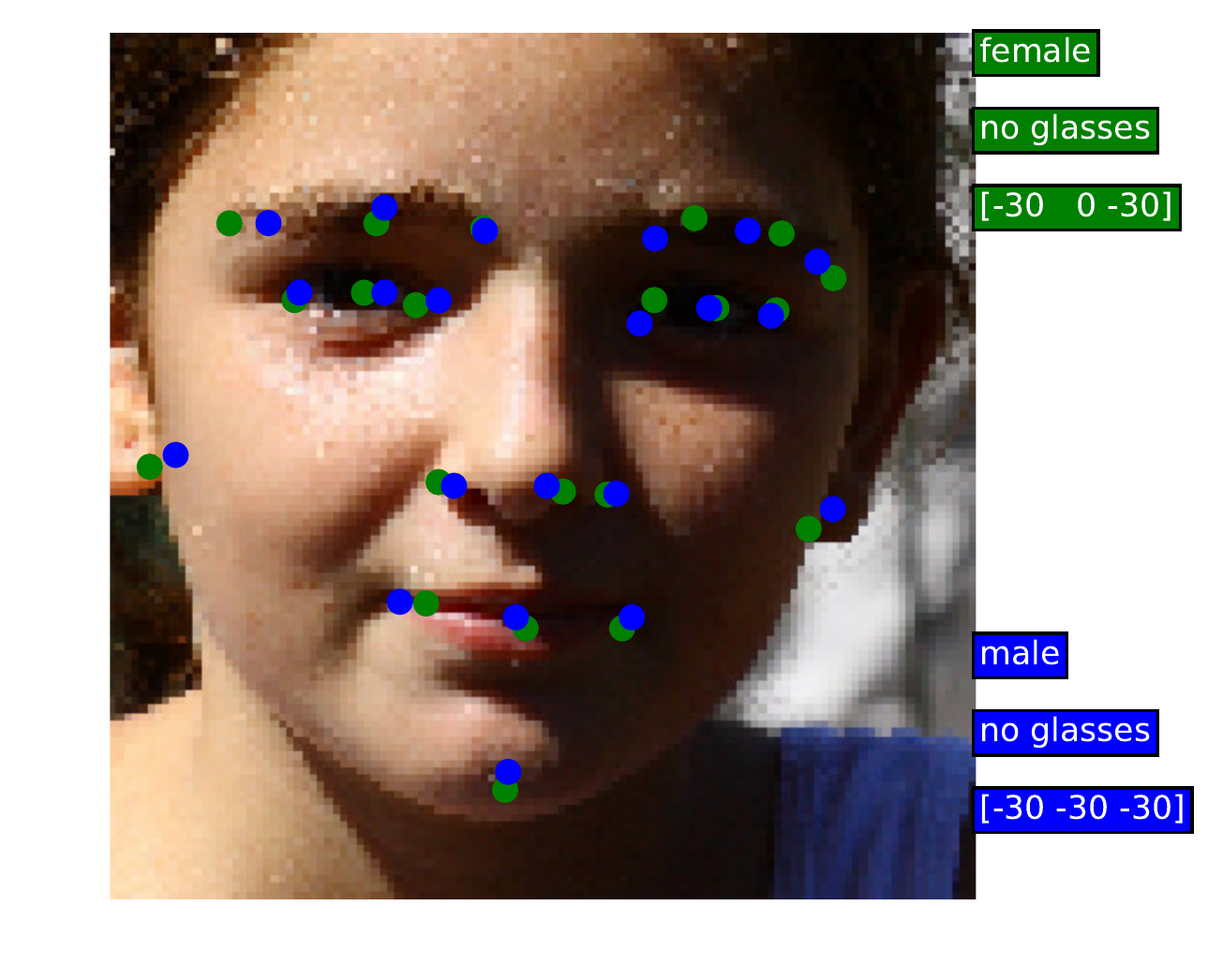}
        \end{minipage}
        \begin{minipage}{0.24\linewidth}
            \includegraphics[width=\linewidth]{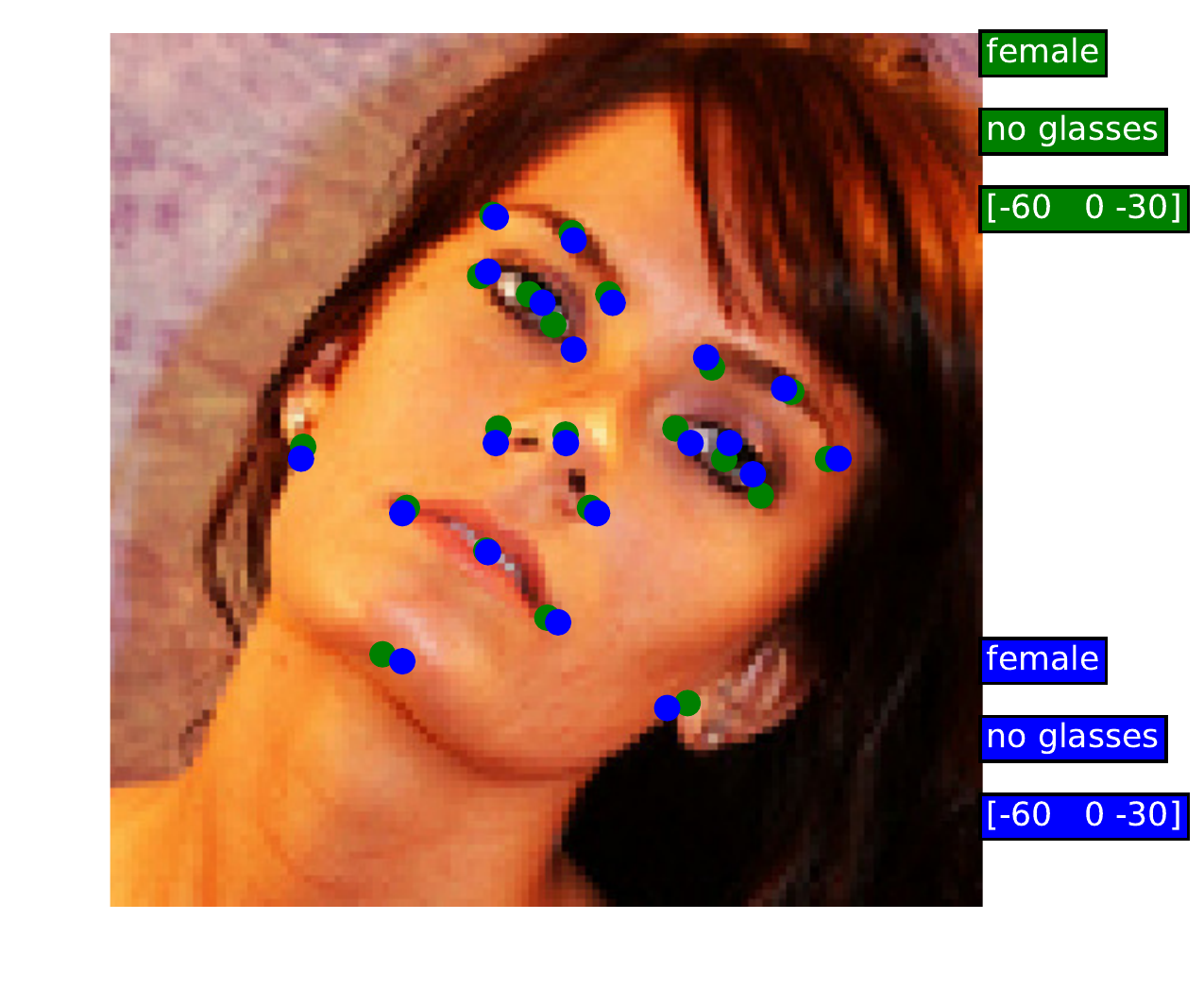}
        \end{minipage}
        \begin{minipage}{0.24\linewidth}
            \includegraphics[width=\linewidth]{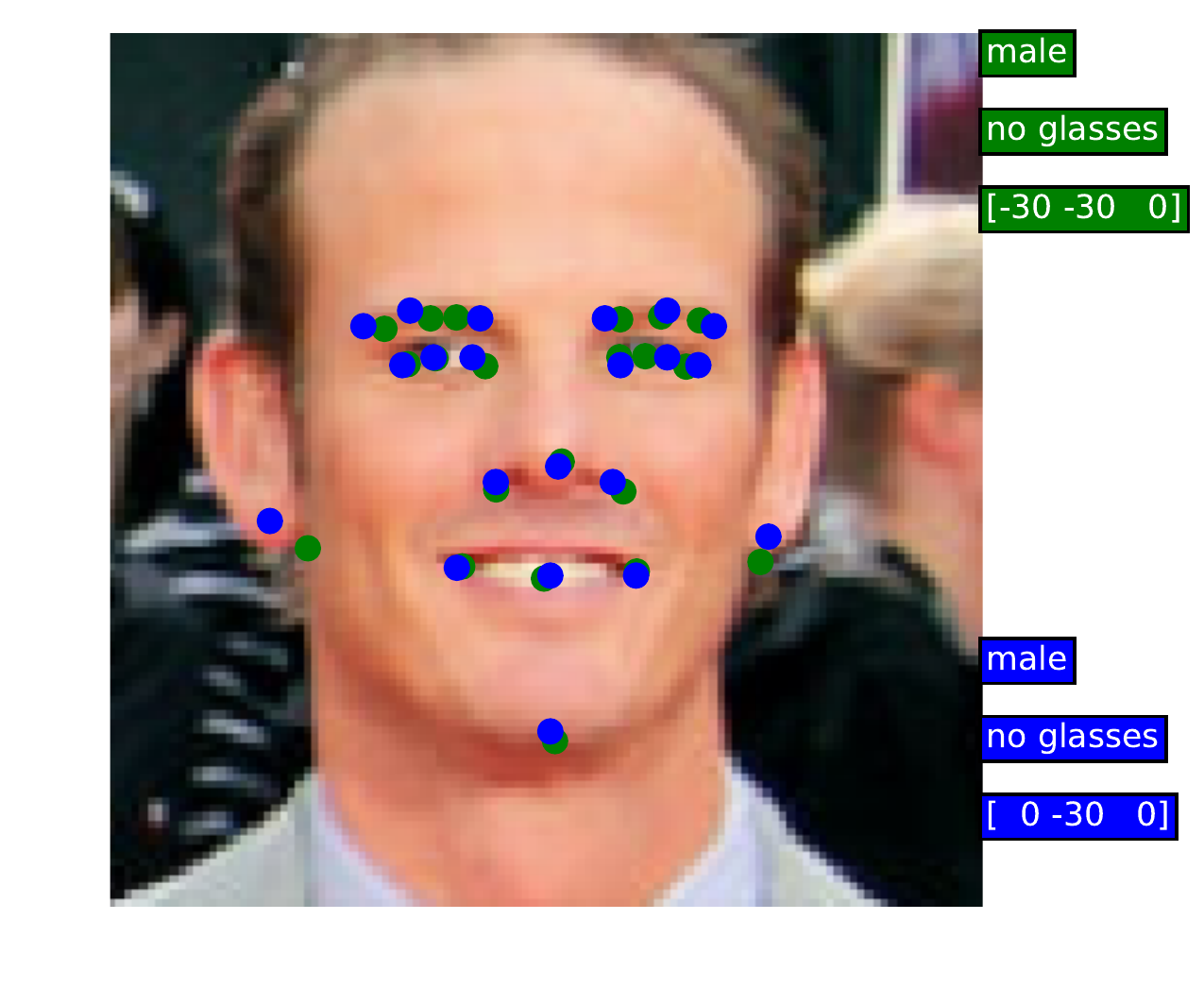}
        \end{minipage}
        \begin{minipage}{0.24\linewidth}
            \includegraphics[width=\linewidth]{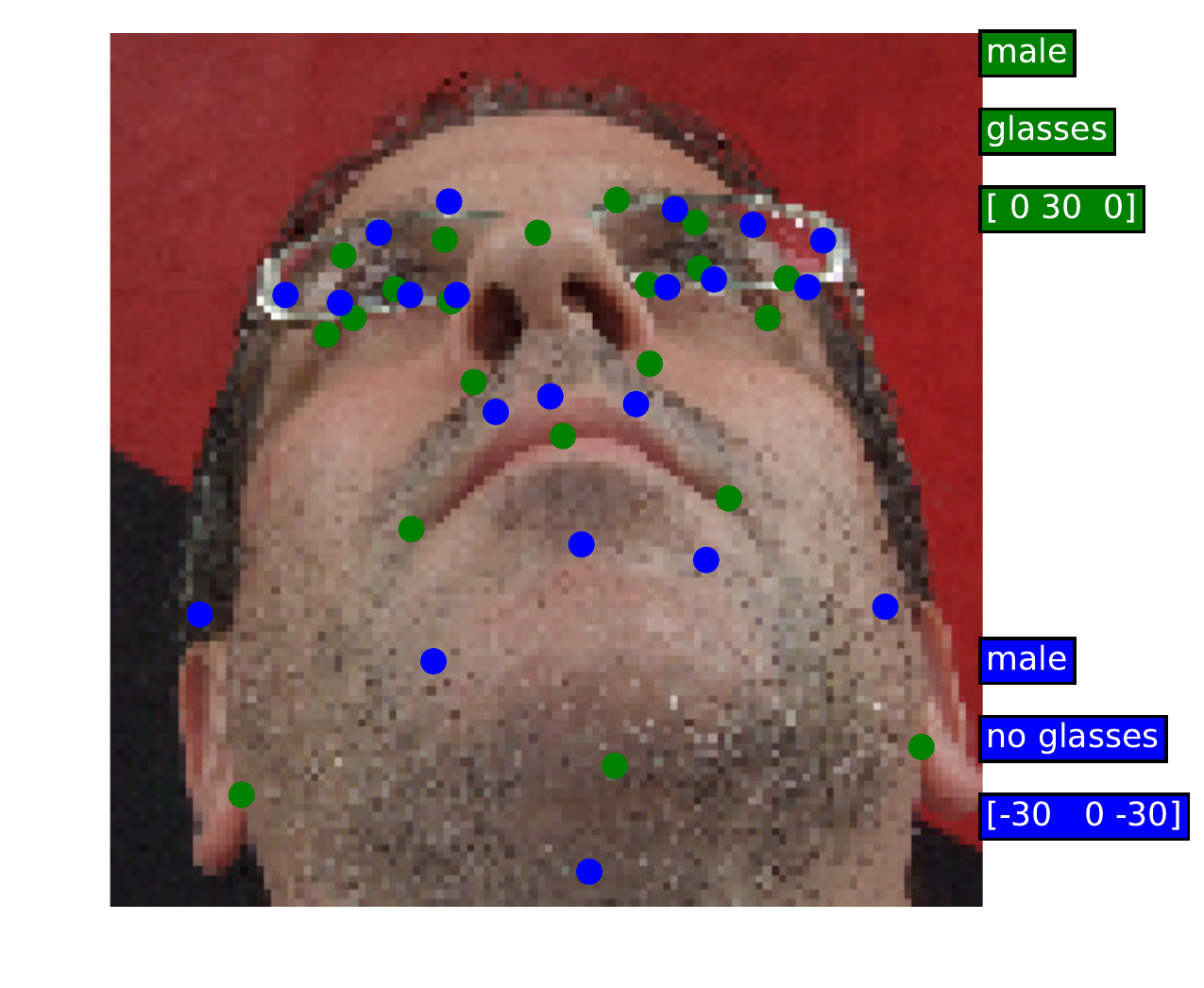}
        \end{minipage}
    \end{minipage}
    \caption{Example predictions of our DCNet with pre-trained ResNet18 as underlying network on the MTFL and AFLW task. For MTFL, the first two examples are successes, while last two are failure cases. For AFLW, the first three examples are successes, while the last one is a failure case. Elements in green correspond to ground truth, while those in blue correspond to predictions. Facial landmarks are shown as small dots, and related tasks labels are displayed on the side. As we can see, over-exposition and tilted face profile can have a large impact on the prediction quality.}
    \label{fig:example-predictions}
\end{figure}

As a first experiment, we performed facial landmark detection on the Multi-Task Facial Landmark (MTFL) task~\citep{zhang2014facial}. The dataset contains 12,995 face images annotated with five facial landmarks and four related attributes of gender, smiling, wearing glasses and face profile (five profiles in total). The training set has 10,000 images, while the test set has 2,995 images. We perform four sets of experiments using an ImageNet pre-trained AlexNet, an ImageNet pre-trained ResNet18, an un-pretrained AlexNet and an un-pretrained ResNet18 as underlying networks. For AlexNet, we apply our collaborative block after each max pooling layer, while for ResNet18, we do as shown in Fig.~\ref{fig:collaborative-resnet18}.

We compare our approach to several other approaches of the literature. We include single-task learning (AN-S when using AlexNet as underlying network, RN-S when using ResNet18), hard-parameter sharing MTL (AN and RN), hard-parameter sharing MTL where the central section is widened to match the number of parameters of our approach (ANx and RNx), HyperFace (HF)~\citep{ranjan2016hyperface}, Tasks-Constrained Deep Convolutional Network (TCDCN)~\citep{zhang2014facial}, Cross-Stitch (XS)~\citep{misra2016cross} and XS widen to match the number of parameters of our approach (XSx). Except for TCDCN, we train each network ourselves three times for 300 epochs and report landmark failure rates averaged over the last five epochs, further averaged over the three tries.

Fig.~\ref{fig:results-mtfl} presents our FLD results on the MTFL dataset. The left part of the figure corresponds to using AlexNet as underlying network, while the right one corresponds to ResNet18. The top part reports the landmark failure rates, while the bottom part reports the mean error. In each plot, the left bar (blue) is for un-pretrained network, while the right bar (green) is for ImageNet pre-trained network. In addition, Fig.~\ref{fig:example-predictions} shows example predictions from DCNet with pre-trained ResNet18 as underlying network. The first two examples were reported as successes, while the last two are failures. The ground truth elements are colored in green, while our predictions are colored in blue. We also include the labels of the related tasks: gender, smiling, wearing glasses and face profile.

The results of Fig.~\ref{fig:results-mtfl} show that our proposed approach obtained the lowest failure rates and mean error in each case. Indeed, our DCNet with un-pretrained and pre-trained AlexNet as underlying network obtained 19.67\% and 19.96\% failure rates respectively, and 14.95\% and 13.52\% with  ResNet18. This is significantly lower than the other approaches to which we compare ourselves. For instance, with AlexNet, HF had 27.75\% and 27.32\%, XS had 26.41\% and 25.65\%, TCDCN had 25.00\%\footnote{Zhang \textit{et. al} only provided results with pre-trained AlexNet~\citep{zhang2014facial}}, and XSx had 25.23\%. With ResNet18, XS had 18.43\% and 15.52\% respectively, and XSx had 17.28. We obtained the highest improvements when using AlexNet as the underlying network when comparing to XS. With un-pretrained and pre-trained AlexNet, we obtained improvements of 6.74\% and 5.69\%, while we obtained 3.48\% and 2.00\% with ResNet18. Performing MTL with our approach can thus improve performance over using other approaches of the literature.

Another result that we can see from Fig.~\ref{fig:results-mtfl} is that our soft-parameter sharing approach obtains higher performance than the hard-parameter sharing approaches with matching number of parameters. For instance, increasing the number of parameters of hard-parameter sharing AlexNet lowers it error rate from 28.02\% (AN) to 26.88\% (ANx), but our approach lowers it further to 19.67\%. Similarly, increasing the number of parameters of hard-parameter sharing ResNet18 lowers it error rate from 20.05\% (RN) to 16.75\% (RNx), but our approach lowers it further to 14.95\%. These results are interesting because they show that while increasing the number of parameters is an effortless avenue to improve performance, it has limitations. Developing novel approaches to enhance network connectivity in a soft-parameter sharing setting seems more rewarding. This may help to motivate new efforts in this avenue to further leverage domain-information of related tasks.

\subsection{Effect of Data Scarcity on the AFLW Task}

\begin{table}[t]
    \renewcommand{\arraystretch}{1}
    \setlength\tabcolsep{15pt}
    \centering
    \caption{Landmark failure rate results on the AFLW dataset using a pre-trained ResNet18 as underlying network. The presented values are averaged over the last five epochs, further averaged over three tries. The first column is the train / test ratio, and the subsequent ones are the networks: single-task ResNet18 (RN-S), multi-task ResNet18 (RN) and Cross-Stitch network (XS). Our approach obtains the best performance in all cases, except the first one where we observe over-fitting.}
    \label{tab:results-aflw-resnet18}
    \begin{tabular}{ccccc}
        \toprule
        \multirow{2}{*}{
            \begin{minipage}[t]{0.25\columnwidth}%
                \centering
                Train / Test Ratio %
        \end{minipage}} & \multicolumn{4}{c}{Networks} \\
        \cmidrule{2-5}
        & RN-S & RN &  XS & Ours \\
        \midrule
        0.1 / 0.9 & \textbf{57.39} & 58.00 & 73.06 & 60.64 \\
        0.3 / 0.7 & 31.84 & 32.00 & 36.24 & \textbf{29.73} \\
        0.5 / 0.5 & 23.41 & 23.31 & 26.02 & \textbf{20.77} \\
        0.7 / 0.3 & 21.47 & 21.92 & 22.37 & \textbf{18.50} \\
        0.9 / 0.1 & 13.03 & 12.80 & 13.51 & \textbf{10.82} \\
        \bottomrule
    \end{tabular}
\end{table}

As second experiment, we evaluated the influence of the number of training examples to simulate data scarcity on the Annotated Facial Landmarks in the Wild (AFLW) task~\citep{koestinger2011annotated}. The dataset has 21,123 Flickr images, and each image can contain more than one face. Instead of using the images as provided, we process them using the available face bounding boxes. We extract all faces with visible landmarks, which gives a total of 2,111 images. This dataset defines 21 facial landmarks and has 3 related tasks (gender, wearing glasses and face orientation). For face orientation, we divide the roll, yaw and pitch into 30 degrees wide bins (14 bins in total), and predict the label corresponding to each real value.

Our experiment works as follows. With a pre-trained ResNet18 as underlying network, we compare our approach to single-task ResNet18 (RN-S), multi-task ResNet18 (RN) and Cross-Stitch network (XS) by training on a varying number of images. We use five different train / test ratios, starting with 0.1 / 0.9 up to 0.9 / 0.1 by 0.2 increment. In other words, we train each approaches on the first $r\%$ of the available images and test on the other $(1-r)\%$, then repeat for all the other train / test ratios. We use the same training framework as in section~\ref{ssec:mtfl-results}. We train each network three times for 300 epochs, and report the landmark failure rate averaged over the last five epochs, further averaged over the three tries. Example predictions are shown in Fig.~\ref{fig:example-predictions}.

As we can see in Table~\ref{tab:results-aflw-resnet18}, our approach obtained the best performance in all cases except the first one. Indeed, we observe between 1.98\% and 6.51\% improvements with train / test ratios from 0.3 / 0.7 to 0.9 / 0.1, while we obtain a negative relative change of 3.25\% with train / test ratio of 0.1 / 0.9. However, since all multi-task approaches obtained higher failure rates than the single-task approach, this suggests that the networks are over-fitting the small training set. Nonetheless, these results show that we can obtain better performance using our approach.

One particularity that we observe in Table~\ref{tab:results-aflw-resnet18} is that the XS network has relatively high failure rates. In the previous experiment of Section~\ref{ssec:mtfl-results}, XS had either similar or better performance than the other approaches (except ours). This could be due to our current multi-task learning framework that is unfavorable towards XS. In order to investigate whether this is the case, we perform the following additional experiment. Using a pre-trained ResNet18 as underlying network, we compare our approach to XS by training each network 100 times using task weights randomly sampled from a log-uniform distribution. Specifically, we first sample from a uniform distribution $\gamma \thicksim \mathcal{U}(\log(1\mathrm{e}{-4}), \log(1))$, then use $\lambda = \exp(\gamma)$ as the weight. We trained both XS and our approach for 300 epochs with the same task weights using a train / test ratio of 0.5 / 0.5.

\begin{figure}[t]
    \centering
    \begin{minipage}{0.49\linewidth}
        \includegraphics[width=\linewidth]{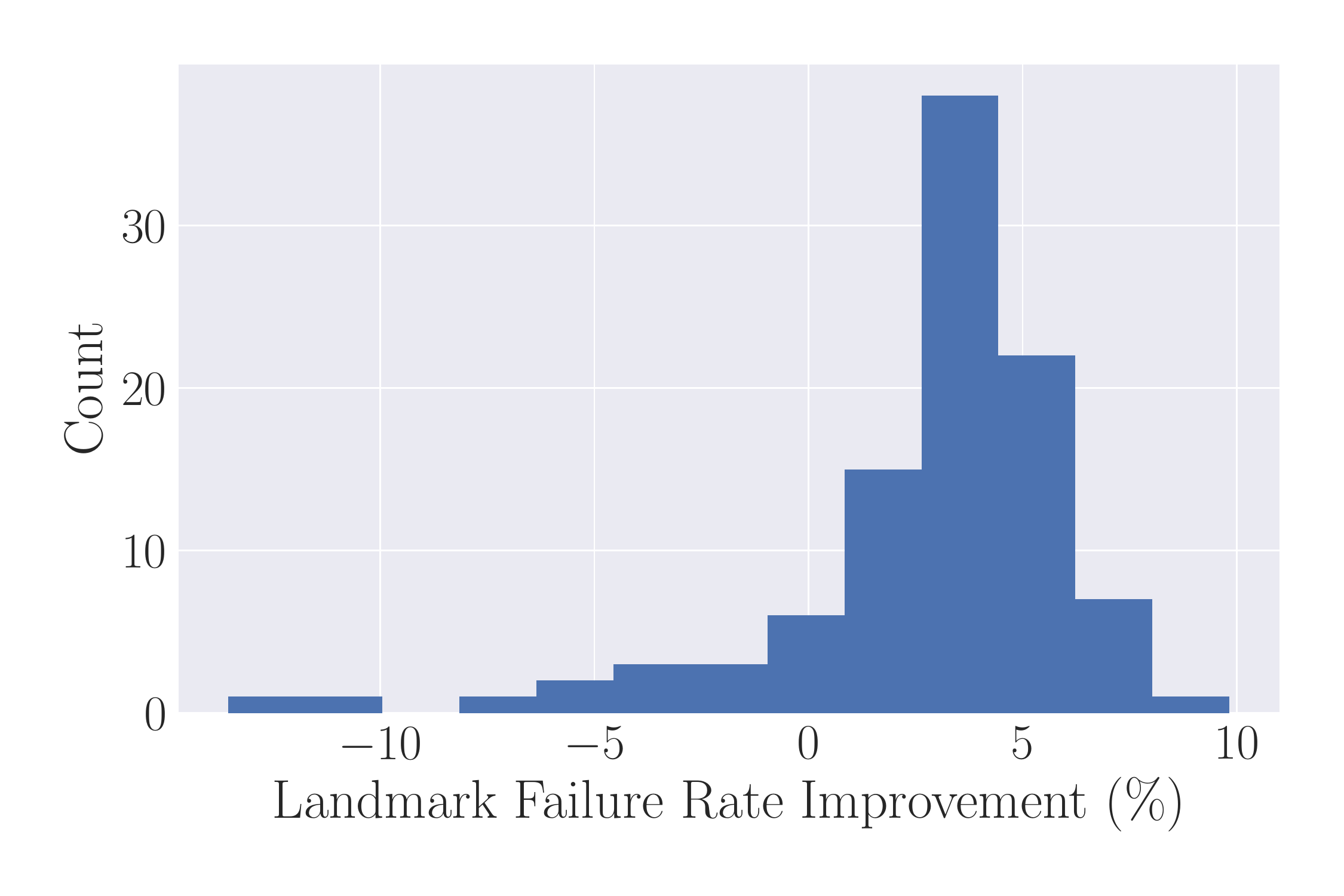}
    \end{minipage}
    \begin{minipage}{0.49\linewidth}
        \includegraphics[width=\linewidth]{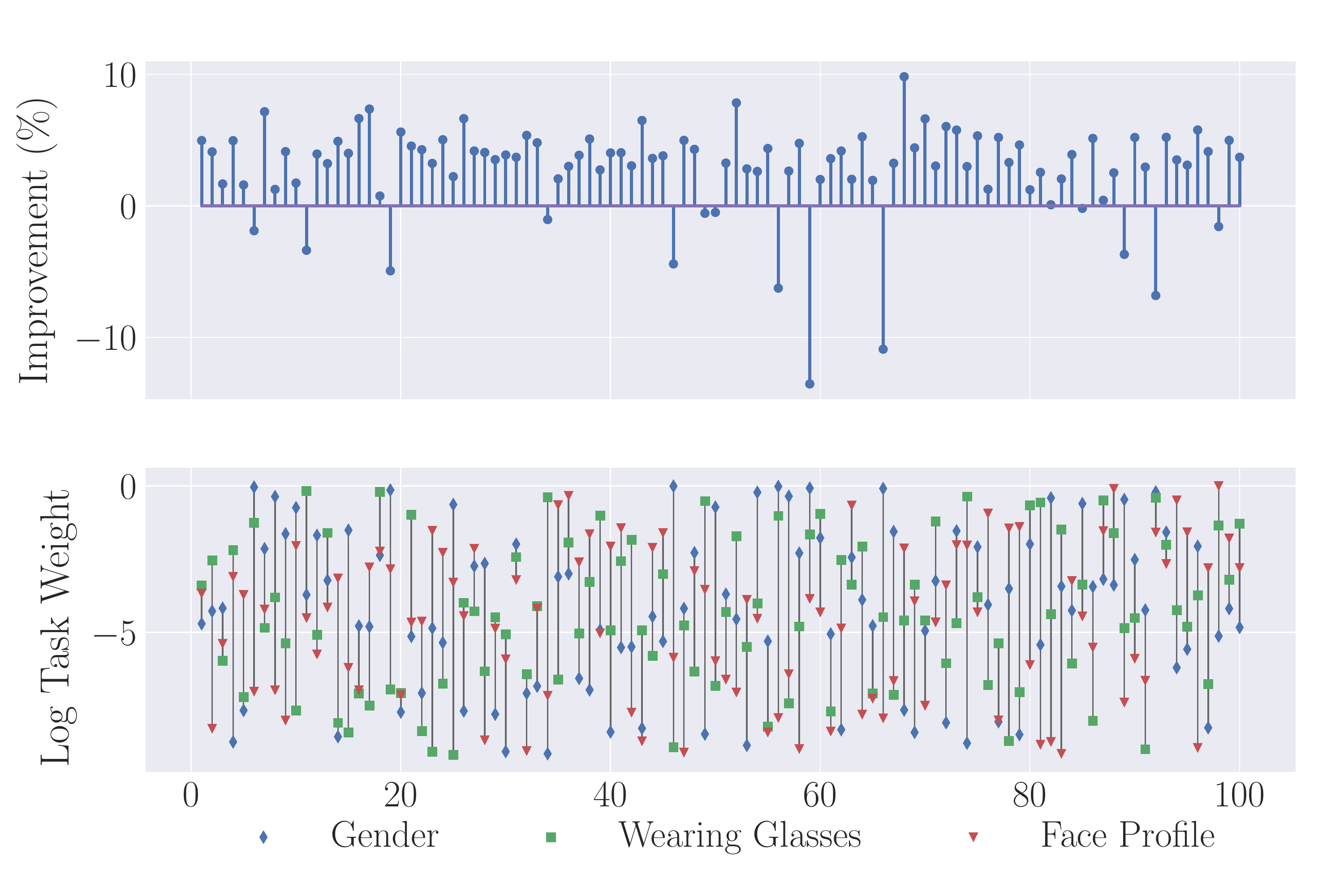}
    \end{minipage}
    \caption{Landmark failure rate improvement (in \%) of our approach compared to XS when sampling random task weights. We used a pre-trained ResNet18 as underlying network. The histogram at the left and the plot at the top right represents performance improvement achieved by our proposed approach (positive value means lower failure rates), while the plot at the bottom right corresponds to the log of the task weights. Our approach outperformed XS in 86 out of the 100 tries, thus empirically demonstrating that our learning framework was not unfavorable towards XS and that our approach is less sensitive to the task weights $\lambda$.}
    \label{fig:aflw-random-task-weights}
\end{figure}

Figure~\ref{fig:aflw-random-task-weights} presents the results of this experiment. The plot at the top right of the figure represents the landmark failure rate improvement (in \%) of our approach compared to XS, while the plot at the bottom right corresponds to the log of the task weights for each try. In 86 out of the 100 tries, our approach had a positive failure rate improvement, that is, obtained lower failure rates than XS. As we can see in the histogram at the left of Fig.~\ref{fig:aflw-random-task-weights}, the improvement rate is normally distributed around 2.80\%, has a median improvement of 3.66\% and a maximum improvement of 9.83\%. Even though we sampled at random the weights of the related tasks, our approach outperforms XS in the majority of the cases. Our learning framework was therefore not unfavorable toward XS.

\subsection{Illustration of Knowledge Sharing With an Ablation Study}
\label{ssec:task-relevance}

As third experiment, we perform an ablation study on the MTFL task~\citep{zhang2014facial} with an un-pretrained ResNet18 as underlying network. The goal of this experiment is to verify that our collaborative block effectively enables knowledge sharing between task-specific CNNs. To do so, we evaluate the impact, on facial landmark detection, of removing the contribution of each task-specific features. We zero out the designated feature map $x_t$ before concatenation at the input of the central aggregation $\mathcal{H}$. The network is trained using the same framework as explained in Sec.~\ref{ssec:mtl-framework}, and the ablation study is performed at test time on the test set when training is done.

\begin{figure}[t]
    \centering
    \includegraphics[width=0.6\linewidth]{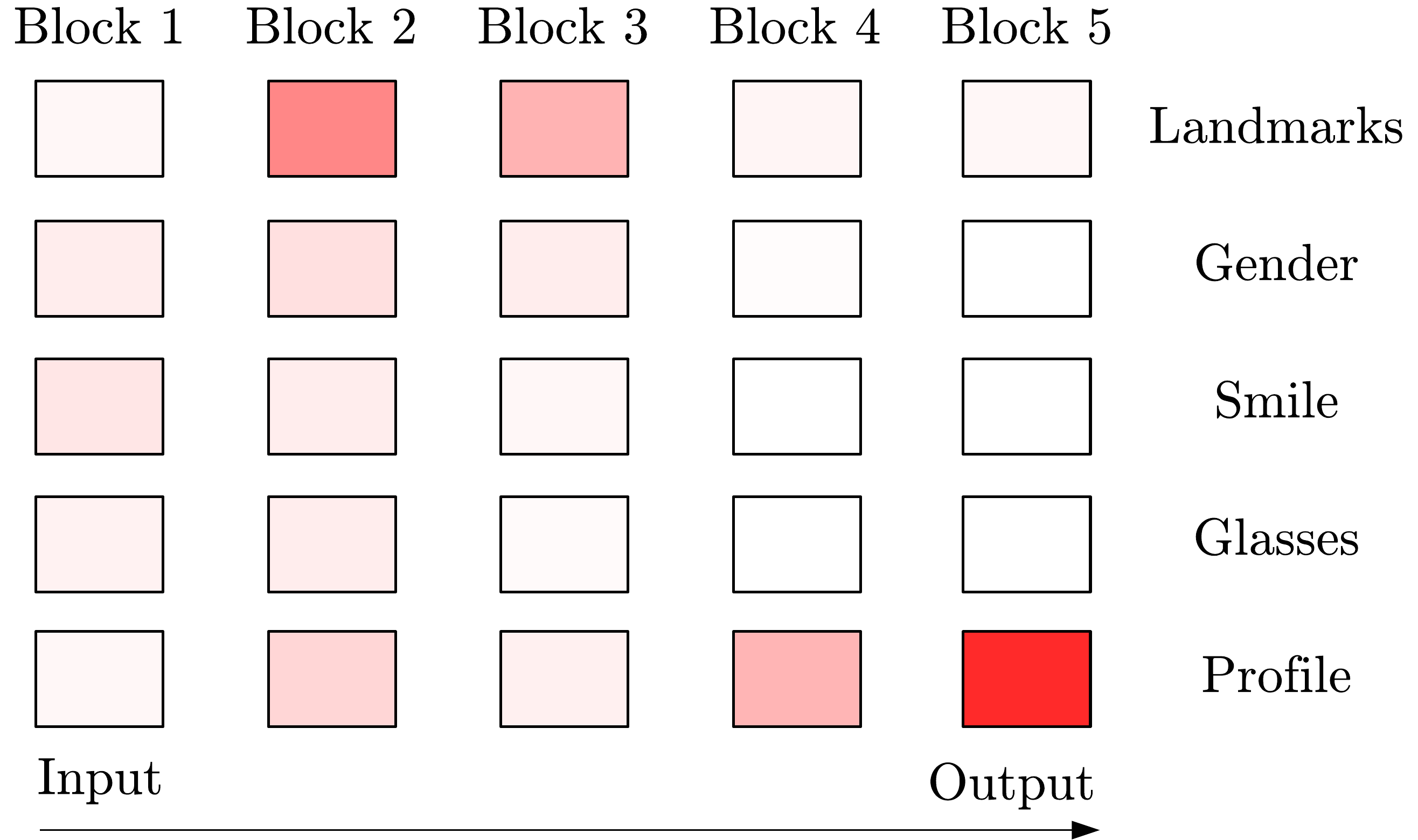}
    \caption{Results of our ablation study on the MTFL dataset with an un-pretrained ResNet18 as underlying network. We remove each task-specific features from each respective central aggregation layer and evaluate the effect on landmark failure rate. The rows represent the task-specific CNNs, while the columns correspond to the network block structure. Blocks with a high saturated color were found to have a large impact on failure rate. In particular, this ablative study shows that the influence of high-level face profile features is large within our proposed architecture. This corroborates with the well-known fact that the location of facial landmarks is closely dependent on the orientation of the face. This constitutes an empirical evidence of domain-specific information sharing via our approach.}
    %
    \label{fig:ablation-results}
\end{figure}

Figure~\ref{fig:ablation-results} presents the results of our ablation study. The rows represent each task-specific CNN, while the columns correspond to the network block structure. The blocks are ordered from left (input) to right (output), while the task-specific networks are ordered from top (main task) to bottom (related tasks). The color saturation indicates the influence of removing the task-specific feature maps from the central aggregation at the corresponding depth. A high saturation reflects high influence on failure rate, while a low saturation reflects low influence. 

As first result, removing features from the facial landmark detection network significantly increases landmark failure rate. For instance, we observe a negative (worse) relative change of 29.72\% and 47.00\% in failure rate by removing features from Block 3 and Block 2 respectively. This illustrates that the main-task network both contributed to and fed from the global features computed by the central aggregation $\mathcal{H}$. The CNN for landmark detection had the possibility to remove the contribution of the global features, and so isolate itself from the other CNNs, but the opposite occurred. We actually observe a mutual influence between the CNNs, where the task-specific features from the facial landmark CNN influence the quality of the global features, which in turn influence the quality of the subsequent task-specific features.

Another result that we can see from Fig.~\ref{fig:ablation-results} is that Block 5 of task Profile has the highest influence on failure rate. We observe a negative relative change of 83.87\% by removing the features maps of task Profile from the central aggregation. What is particularly interesting in this case is that we observe this high relative change at Block 5, which corresponds to the highest block in the network. Since the block lies at the top of the network, it outputs features with a high level of abstraction. We therefore expect that these features represent high-level factors of variation corresponding to face orientation, which should look like a rotation matrix. It therefore makes sense that features representing the orientation of the face would be useful to predict facial landmarks, since we know that the location of the facial landmarks is closely dependent on the orientation of the face. The landmark CNN can use these rich features to better rotate the predicted facial landmarks. This is indeed what we observe in Fig.~\ref{fig:ablation-results}. These results constitute an empirical evidence that our approach allows leveraging domain-specific information from related tasks.

\subsection{Facial Landmark Detection With MTCNN}
\label{ssec:mtcnn}

As final experiment, we performed an experimental evaluation using the recent Multi-Task Cascaded Convolutional Network (MTCNN)~\citep{zhang2016joint}. The authors proposed a cascade structure of three stages, where each stage is composed of a multi-task CNN. MTCNN performs predictions in a coarse-to-fine manner. The CNN of the first stage generates (in a fully-convolutional way) many hypotheses about the position of the face and the facial landmarks, and the subsequent second and third stages refines them. The CNNs are trained sequentially with hard-negative mining, in a hard-parameter sharing setting.

We implemented our approach in the available code project~\citep{kuaikuaikim2018DFace} and compared ourself to MTCNN. We followed the provided hard negative mining recipe and generated our images. For landmark detection, we used the LFWNet~\citep{sun2013deep} and CelebA~\citep{liu2015faceattributes} datasets, and generated 600k face images with facial landmarks. For face detection, we used the WIDER~\citep{yang2016wider} dataset, and generated 1.5M face images with a bounding box. We trained a MTCNN with the stage networks connected with our collaborative block, and a standard MTCNN with widened stage networks to match the number of parameters.

\begin{figure}[t]
    \centering
    \includegraphics[width=\linewidth]{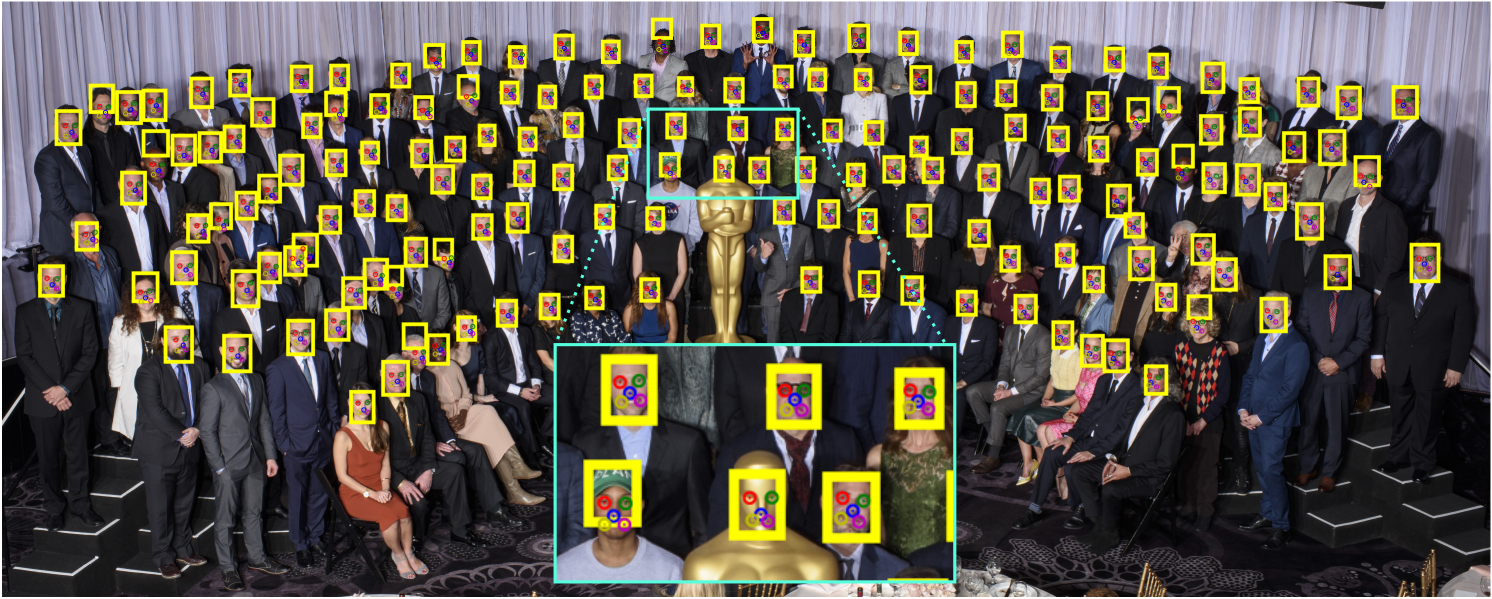}
    \caption{MTCNN predictions on the photo of the 2018 Oscar nominees (image resolution of 2983 $\times$ 1197). The stage-CNNs are trained using our proposed collaborative block. The coarse-to-fine detection scheme employed by MTCNN allows predicting many faces in high-dimensional images with low computational burden.}
    \label{fig:oscar-2017predcollaborative}
\end{figure}

On the test set of MTFL~\citep{zhang2014facial} dataset, standard MTCNN obtained a landmark failure rate of 37.85\%, a mean error of 0.0996 and failed to detect a face 112 times, while our approach obtained better performances with 28.97\%, 0.0930 and 79 respectively. Note that the reason MTCNN obtains worse performance than our Deep Collaboration Network (DCNet), as reported in Fig.~\ref{fig:results-mtfl}, is because it has fewer parameters. DCNet has about 85M parameters, while the sum of all three stage-CNN in MTCNN is about 2M. This is because MTCNN is carefully designed to balance computational speed and landmark detection precision. It can predict many faces in high-dimensional images with a low computation burden. An example of its prediction capability is shown in Fig.~\ref{fig:oscar-2017predcollaborative}.

\section{Conclusion and Future Work}
\label{sec:conclusion}
In this paper, we proposed a novel soft-parameter knowledge sharing mechanism based on lateral connections for Multi-Task Learning (MTL). Our proposed approach implements connectivity in term of a \textit{collaborative block}, which uses two distinct non-linear transformations. The first one aggregates task-specific features into global features, and the other merges back the global features into each task-specific Convolutional Neural Network (CNN). Our collaborative block is differentiable and can be dropped in any existing CNN architectures as a whole.
Our results on facial landmark detection tasks showed that networks connected with our proposed collaborative block outperformed the other state-of-the-art approaches, including the recent Cross-Stitch and MTCNN approach. We verify that our collaborative block effectively enables knowledge sharing between task-specific CNNs with an ablation study. We observed that the CNNs incorporated features with a varying level of abstraction from the other CNNs, by observing the depth-specific influence of tasks that we know are related. These results constituted an empirical evidence that our approach allows leveraging domain-specific information from related tasks. 
Evaluating our proposed approach on other MTL problems could be an interesting avenue for future works. For instance, the recurrent networks used to solve natural language processing problems could benefit from our approach.

\section*{Acknowledgements}

We gratefully acknowledge the support of NVIDIA Corporation for providing a Tesla Titan X for our experiments through their Hardware Grant Program.

\bibliographystyle{splncs}
\bibliography{egbib}

\end{document}